\documentclass[conference]{IEEEtran}
\IEEEoverridecommandlockouts
\usepackage{cite}
\usepackage{amsmath,amssymb,amsfonts}
\usepackage{algorithmic}
\usepackage{graphicx}
\usepackage{textcomp}
\usepackage{xcolor}
\usepackage{multirow}
\def\BibTeX{{\rm B\kern-.05em{\sc i\kern-.025em b}\kern-.08em
    T\kern-.1667em\lower.7ex\hbox{E}\kern-.125emX}}

\begin{document}

\title{An Integrated Framework for Team Formation and Winner Prediction in the FIRST Robotics Competition: Model, Algorithm, and Analysis}

\makeatletter
\newcommand{\linebreakand}{%
  \end{@IEEEauthorhalign}
  \hfill\mbox{}\par
  \mbox{}\hfill\begin{@IEEEauthorhalign}
}
\def\ps@IEEEtitlepagestyle{
  \def\@oddfoot{\mycopyrightnotice}
  \def\@evenfoot{}
}
\def\mycopyrightnotice{
  {\footnotesize
  \begin{minipage}{\textwidth}
  \centering
  \copyright~2023 IEEE.  Personal use of this material is permitted.  Permission from IEEE must be obtained for all other uses, in any current or future media, including reprinting/republishing this material for advertising or promotional purposes, creating new collective works, for resale or redistribution to servers or lists, or reuse of any copyrighted component of this work in other works.
  \end{minipage}
  }
}

\makeatother

\author{\IEEEauthorblockN{Federico Galbiati}
\IEEEauthorblockA{\textit{Department of Computer Science} \\
\textit{Worcester Polytechnic Institute}\\
Worcester, USA \\
fgalbiati@wpi.edu}
\and
\IEEEauthorblockN{Ranier X. Gran}
\IEEEauthorblockA{\textit{Data Science Program} \\
\textit{Worcester Polytechnic Institute}\\
Worcester, USA \\
rxgran@wpi.edu}
\and
\IEEEauthorblockN{Brendan D. Jacques}
\IEEEauthorblockA{\textit{Data Science Program} \\
\textit{Worcester Polytechnic Institute}\\
Worcester, USA \\
bdjacques@wpi.edu}
\linebreakand
\IEEEauthorblockN{Sullivan J. Mulhern}
\IEEEauthorblockA{\textit{Department of Computer Science} \\
\textit{Worcester Polytechnic Institute}\\
Worcester, USA \\
sjmulhern@wpi.edu}
\and
\IEEEauthorblockN{Chun-Kit Ngan}
\IEEEauthorblockA{\textit{Data Science Program} \\
\textit{Worcester Polytechnic Institute}\\
Worcester, USA \\
cngan@wpi.edu}
}

\maketitle

\begin{abstract}
This research work aims to develop an analytical approach for optimizing team formation and predicting team performance in a competitive environment based on data on the competitors’ skills prior to the team formation. There are several approaches in scientific literature to optimize and predict a team’s performance. However, most studies employ fine-grained skill statistics of the individual members or constraints such as teams with a set group of members. Currently, no research tackles the highly constrained domain of the FIRST Robotics Competition. This research effort aims to fill this gap by providing an analytical method for optimizing and predicting team performance in a competitive environment while allowing these constraints and only using metrics on previous team performance, not on each individual member's performance. We apply our method to the drafting process of the FIRST Robotics competition, a domain in which the skills change year-over-year, team members change throughout the season, each match only has a superficial set of statistics, and alliance formation is key to competitive success. First, we develop a method that could extrapolate individual members' performance based on overall team performance. An alliance optimization algorithm is developed to optimize team formation and a deep neural network model is trained to predict the winning team, both using highly post-processed real-world data. Our method is able to successfully extract individual members' metrics from overall team statistics, form competitive teams, and predict the winning team with 84.08\% accuracy.
\end{abstract}


\section{Introduction}
\label{sec:introduction}
The best functioning team is a team that is better than the sum of its individuals. While a team may benefit from having individual members, who possess various useful skills to support the team in reaching the final goal, it is still difficult to determine the contributions of each one of them due to the existence of a group dynamic, especially if the information on each individual member's performance is missing or imprecise. Thus, this work addresses two critical challenges: (1) forming an effective team based on the strengths and weaknesses of its individual members and (2) predicting the winning team participating in a competition.

For us to address these two challenges, in this research work, we mainly focus on the domain of the ``For Inspiration and Recognition of Science and Technology'' (FIRST) Robotics Competition (FRC). The FRC is a team-focused tournament, where each participating team needs to build a robot to take on the specialized challenges. During the competition, each participating team's robot joins together to form a 3-robot ``alliance'' to match their complementary skills and compete in a tournament against other alliances. Due to the presence of various domain constraints, FRC makes itself become a unique problem space: (1) the team-making model must maximize the competitiveness of a team, (2) a flat hierarchy dominates the teams, (3) the performance assessment in the competition can only be captured at the alliance level but not at the team level, (4) diversity of skills is essential, (5) no baseline pre-test data of a team can be collected, and (6) the opponent alliance cannot be predefined.

To the best of our knowledge, no previous work on team formation and winner prediction can be found in the scientific literature for our domain problem and more generally with this set of constraints. Thus, this research work aims to provide a feasible solution to address the above challenges. Specifically, we propose an integrated framework that is a data-driven processing pipeline to power the team formation and the winner prediction. Our strategy assists teams with the selection of alliance members by considering their skills and predicting their chances of winning. Data of the qualification matches, i.e., competitions needed to qualify for the playoffs, is firstly used to evaluate how randomized alliances in the tournament performed. Each alliance is evaluated using seven criteria: (1) Traditional Scoring Low, (2) Traditional Scoring High, (3) Technical Scoring, (4) Autonomous Scoring, (5) Endgame Scoring, (6) Fouls, and (7) Defense. We evaluate the quality of our alliance optimization algorithm using our winner prediction model to simulate a match and predict its outcome with a model with 84.08\% accuracy.

The rest of the paper is organized as follows. In Section~\ref{sec:related-work}, we discuss the scientific literature for attempts to solve the related problems. Section~\ref{sec:methodology} is our proposed framework that includes an overview of the competition as well as the method for data collection and pre-processing, our alliance optimization algorithm, and the deep learning model training, testing, and evaluation. Section~\ref{sec:results} provides and discusses the experimental results achieved using the framework described in Section~\ref{sec:methodology}. Lastly, Section~\ref{sec:conclusions} includes the conclusions and suggestions for our future work.

\section{Related Work}
\label{sec:related-work}
Our research explores the team formation challenge and then draws inspiration from the field of automated personnel drafts using machine learning solutions. The team formation problem is a quickly growing space that informally refers to finding team members to maximize the team's effectiveness. Several works identify different conceptions of team formulation problems based on their specific domains \cite{juarezComprehensiveReviewTaxonomy2021}. In the context of team formation, there are several studies on the factors that are considered in the algorithmic space and the actual optimization solutions. The trade-offs involved in team effectiveness, computational time, and policy considerations must all be considered \cite{sidhuGeneralizedFrameworkAlgorithm2021}. The task of evaluating teamwork skills is a significant problem on its own \cite{kotlyarAssessingTeamworkSkills2022}.

The scientific literature suggests classifying matchmaking systems into the following categories: random, quasi-random, skill-based, role-based, technical factor-based, and engagement-based. The most common systems used in eSport games are skill and rank-based systems \cite{marcinAnalysisMatchmakingOptimization2019}. There are several examples of optimizing the drafting process on the scale of individual sports games \cite{summervilleDraftAnalysisAncientsPredicting2016, wardAISolutionsDrafting2021}, predicting the future performance of individual athletes and competitors based on specific factors \cite{muazumusaApplicationArtificialNeural2019}, predicting the performance of different group configurations in completing long-term tasks \cite{nikolakakiCompetitiveBalanceTeam2020, giannakasDeepLearningClassification2021}, and optimizing the drafting process on a season-to-season scale \cite{alcoxApplicationsArtificialIntelligence2019}.

A large body of research on team formation only considers the team itself in isolation from the competitive environment. A study has involved multi-stage stochastic algorithms for team formation, but only to optimize two properties: cost and competency \cite{rahmanniyayMultiobjectiveMultistageStochastic2019}. Others have focused on optimizing team formation while minimizing fault lines within the team, but without trying to maximize the competitiveness \cite{bahargamTeamformationAlgorithmFaultline2019}. Others also focus on optimizing teams based on the generally accepted nine Belbin team roles or the experts and laymen problem \cite{ugarteUsingBehaviouralTendency2022, yehGreedyBasedPreciseExpansion2022}. Such an approach is not generalizable in a domain where a flat hierarchy dominates the teams.

Algorithms are developed for team formation in the field of software development, often using bio-inspired algorithms such as genetic algorithms \cite{costaTeamFormationSoftware2020, ahmadAdvanceRecommendationSystem2019}. However, these often require fine-grained data for each member, using data mining techniques or manual intervention for the dataset creation \cite{ahmadAdvanceRecommendationSystem2019, kouEfficientTeamFormation2020, shiChallengesOpportunitiesDataCentric2021, afsharFindingTeamSkilled2020, andrejczukSynergisticTeamComposition2019, gomez-zaraSearchDiverseConnected2022, kalantziFERNFairTeam2022, abidinCaseStudyPlayer2021, rajeshDataScienceApproach2020, silvaAllNBATeamsSelection2022}. Such precise data is not available in our domain where performance metrics are captured for the overall team as opposed to fine-grained for each member.

Others research the domains where the in-depth competence for a general set of skills is fundamental, but the diversity of skill types does not contribute to better performance \cite{sidhuEffectivenessFactorsAlgorithm2021}. Such team-making approaches are inapplicable in our domain, where diversity in background and skills is fundamental for the overall success of the team.

Many team-making domains allow acquiring baseline pre-test data of a member or surveys to help form teams \cite{gavrilovicDesignImplementationDiscrete2022, liangLearningLogbasedAutomatic2021}. This is inadequate in our domain where members of teams change throughout a season, and there is no baseline score or survey taken in our domain.

Others use genetic algorithms to optimize group formation to promote collaborative problem-based learning but by balancing teams so that members have similar skill levels \cite{chenOptimizedGroupFormation2019}, which is inapplicable to our competitive domain, where each team member often has a specialized role described by skill metrics.

Team formation is also fundamental in the space of sports. Algorithms are developed to optimize based on only the adversary teams participating in the competition \cite{vetukuriGenericModelAutomated2021}. In our domain, the opponent team is not predefined, which is an additional constraint that makes several studies inapplicable.

Various methods are proposed to address our second challenge of winner prediction. A recent paper proposes a framework coupled with a prediction model \cite{wangMatchTracingUnified2020}. This framework performs tracking of the win prediction and performance evaluation jointly. This provides a win-rate curve over time, instead of a binary prediction. However, this method requires very fine-grained information about the match, which is unavailable in our domain.

Another paper proposing the Winning Tracker framework similarly relies on detailed data including individual movement and confrontation information \cite{zhaoWinningTrackerNew2022}. Other research suggests performing winner prediction using detailed time series data \cite{yangModifiedMultisizeConvolution2019}. Although there are previous projects to perform action recognition and extraction of features, match videos are often unavailable in our domain making this unfeasible \cite{zhuTransformerbasedSystemAction2022}.

Another recent paper approaches the winner prediction in the game ``CS:GO'' by using the score of each round, the economy of each team, and the map of the match as predictive variables. This approach allows improved predictions as a match progresses by using the results of previous rounds to better predict future rounds \cite{brewerCreatingWellCalibrated2022}. This approach is unfeasible for our domain, where the qualifying matches of a tournament feature randomized teams in each match.

Lastly, one paper uses a deep learning model to predict match outcomes in the game ``League of Legends'' based on the recorded experience level of each player with the character that they are playing. This includes player-champion win rate, the total number of games played on the champion per season, the number of recent games played on the champion, and an internal mastery metric \cite{doUsingMachineLearning2021}. This approach inspires our approach with changes and expansions for our domain.

\section{Domain Overview and Proposed Framework}
\label{sec:methodology}
\subsection{Research Domain}
Each year, the FIRST Robotics Competition (FRC) releases a new robotics game, in which teams compete among themselves to design a robot that specifically accomplishes that game's tasks. We choose to focus on this domain as our research due to the availability of data, the importance of the competition, and the lack of existing research that tackles all the constraints.

\subsection{Dataset Description}
FRC maintains the data from competitions including match data, match recordings, and other services for evaluation and assessment of other teams during competition play. Various APIs exist for data access. For this research, The Blue Alliance (TBA) API is chosen for its ease of use.

Competitions are provided as a set of JSON objects representing individual matches. Each alliance is formed of three teams, i.e., each team provides a robot that can be specialized to perform competition tasks. Although the competition tasks change every year, some elements appear in a similar form year-to-year, particularly the different scoring methods. These scoring methods can be divided into (1) autonomous points, scored by the robot without human interaction, (2) teleoperated points, scored by the robots controlled by a driver, (3) endgame points, scored by completing specific objectives at the end of the match, and (4) foul points scored through penalties given to the opposing team.

Since each season has the same base components, there are shared keys in the match data between years. Two types of data are acquired for each match: (1) Identifiers: Keys are used to identify the match number, event, and participating teams; and (2) Scoring and Match Replays: Keys are used to keep track of who wins the match, how they win the match, and official video recordings of the match.

Testing for this methodology is performed with FRC match data from the 2017, 2018, and 2019 seasons due to their high data availability, as well as the number of events and data points that were significantly impacted in 2020 and 2021 due to the COVID-19 pandemic \cite{BlueAllianceInsights}.

\subsection{Framework Overview}
\label{sec:fwk-overview}

\begin{figure*}[htbp]
\centerline{\includegraphics[width=0.7\linewidth]{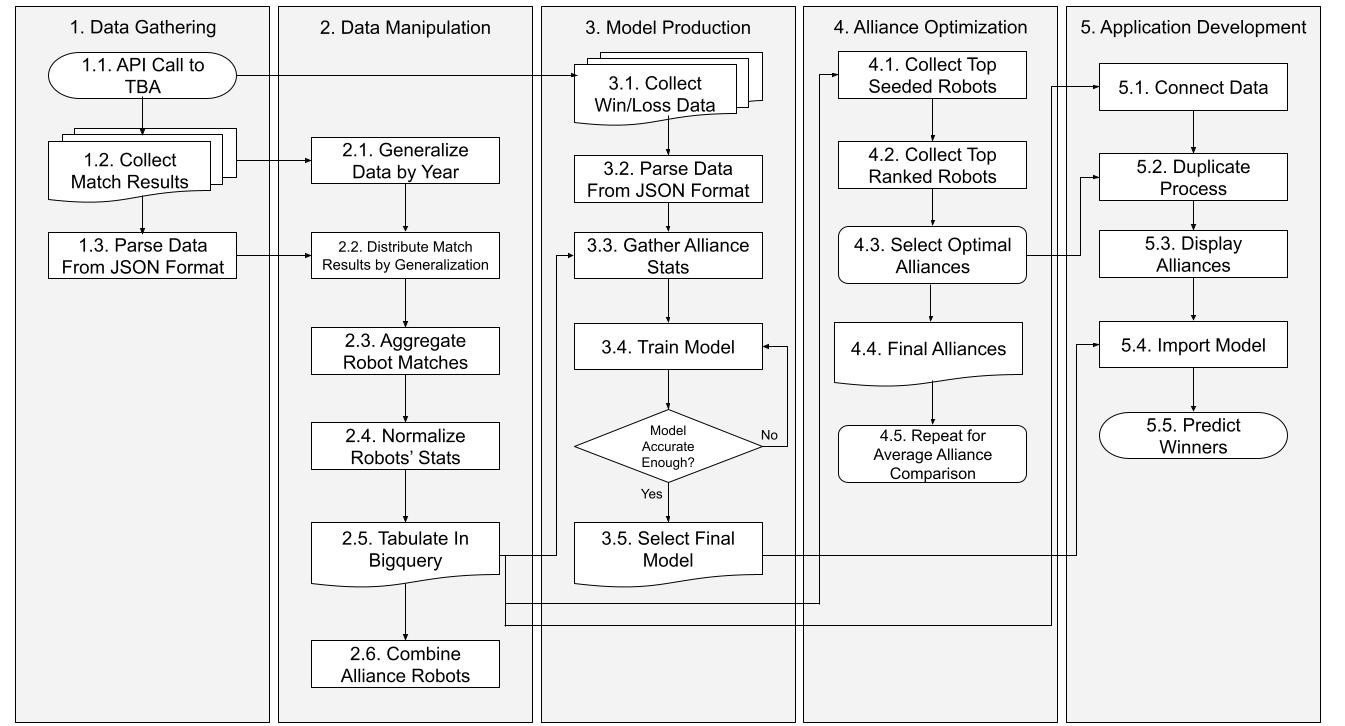}}
\caption{Flow Chart of Our Integrated Framework: A Data-Driven Processing Pipeline to Power the Alliance Optimization Algorithm and the Winner Prediction Model. The Framework Consists of Five Different Modules: (1) Data Gathering, (2) Data Manipulation, (3) Winner Prediction Model Production, (4) Alliance Optimization, and (5) Application Development.}
\label{fig:flow-diagram}
\end{figure*}

Fig.~\ref{fig:flow-diagram} visually explains the steps of our Integrated Framework, which are divided into five modules:
\begin{enumerate}
    \item Data Gathering: The results of every qualification match and the list of robots involved in a given year are parsed from TBA's API, imported into Pandas, and used for (2) Data Manipulation and (3) Winner Prediction Model Production.
    \item Data Manipulation (Section~\ref{sec:data-manipulation}): Seven performance attributes are created to evaluate the strengths and weaknesses of each alliance in a match. These evaluators, together with the pre-processing steps, allow extracting normalized performance of individual robots in the competition instead of alliance-wide statistics.
    \item Winner Prediction Model Production (Section~\ref{sec:model-production}): The qualifying match data and the performance indicators of a given year from (2) Data Manipulation are used to predict the winner of any given match using a multi-layer perceptron.
    \item Alliance Optimization Strategy (Section~\ref{sec:alliance-optimization}): Robot evaluations gathered in (2) Data Manipulation are used to generate initial optimal alliances for the top eight teams in the bracket for a given year. An optimal alliance is defined as one with maximal effectiveness in all seven performance criteria given the remaining pool of competing robots.
    \item Application Development: The winner prediction model along with the information gleaned from (2) Data Manipulation and (4) Alliance Optimization are used to power a front-end web application. This application takes the current make-up of the top eight alliances as input and then calculates which of the remaining recruitable robots will lead to the greatest increase in alliance performance, using the winner prediction model to predict the effectiveness of prospective alliances against calculated ``average'' alliances.
\end{enumerate}

\subsection{Data Manipulation}
\label{sec:data-manipulation}
The first processing step of our Data Manipulation module is the development of a schema to compare the strengths and weaknesses of robots and alliances against each other. While it is true that the exact rules of how the competition being judged is changed every year (e.g., how the points are awarded, what challenges are required, etc.), certain elements are still consistent and appropriate among the seven performance indicators: (1) Traditional Scoring Low - A scoring activity that requires simple but consistent actions on the part of the robot, such as tossing balls into a goal or collecting items from a kiosk; (2) Traditional Scoring High - A harder difficulty version of the default scoring activity that is worth more points, such as tossing balls a much further distance or into a smaller target; (3) Technical Scoring - A second scoring activity worth significantly more points, but requires a distinct, more technically difficult skill to complete, such as putting a puzzle piece into an uneven surface; (4) Autonomous Scoring - A period at the beginning of the match where all robots must perform the traditional and technical tasks without human input for increased points; (5) Endgame Scoring - A final goal that earns the alliance bonus points if they can complete it before time expires, such as climbing up a tube structure; (6) Fouls - Bonus points are awarded to an alliance when a robot in the opposing alliance breaks a match rule; and (7) Defense - An action to disrupt the task completion of an opposing robot or alliance such as bumping into them as they attempt to shoot a ball into a hoop so that they miss. Each year's activities are manually aligned with these seven indicators to provide common statistics throughout the years.

For example, the 2019 game requires robots to score using two distinct game pieces: a ball (cargo) and plastic disks (hatch panels). As it is generally easier for a robot to handle a ball than a disk, the ball scoring is linked to Traditional Scoring, while the disks are assigned to Technical Scoring. Since the rocket can be scored at various heights, the rocket scoring is chosen to be the Traditional Scoring High option while the cargo ship is chosen to be the Traditional Scoring Low option. Using this information, as well as the assigned scoring values from FRC for each action, the generalized schema shown in Fig.~\ref{fig:generalized-schema} is formed. This process is repeated for the 2017 and 2018 competitions to achieve a generalized schema, for all three years. The overall processing pipeline is shown in Fig.~\ref{fig:robot_processing_pipe}.

\begin{figure}[htbp]
\centerline{\includegraphics[width=\linewidth]{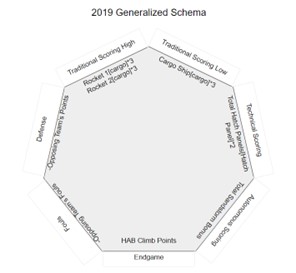}}
\caption{The Generalized Schema for the 2019 FRC Game: The Seven Skill Metrics Matched With Their Original Data Source and Mathematical Relation.}
\label{fig:generalized-schema}
\end{figure}

\begin{figure*}[htbp]
\centerline{\includegraphics[width=0.7\linewidth]{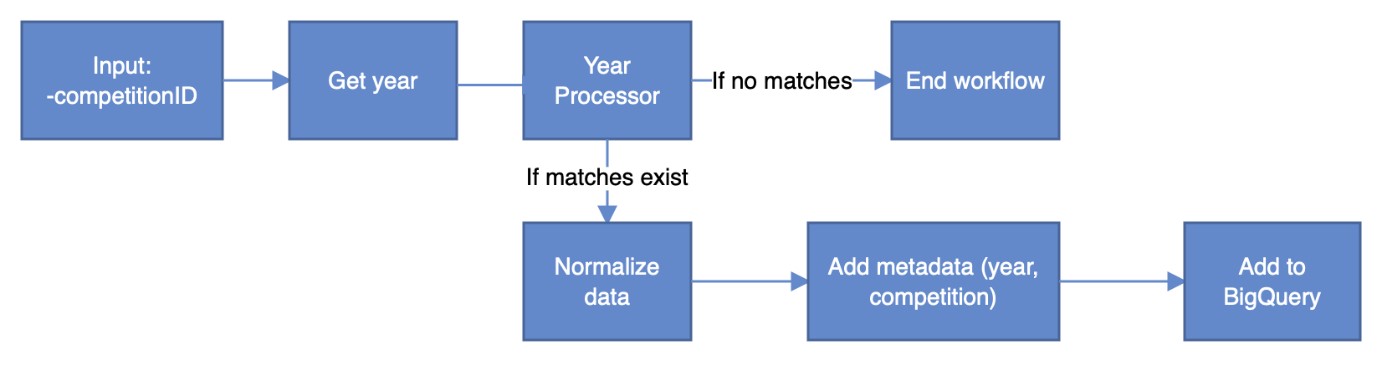}}
\caption{Data Processing Pipeline for the Data Manipulation Phase.}
\label{fig:robot_processing_pipe}
\end{figure*}

As the data coming from TBA's API is specific to each year's competition's game rules and not directly robot-specific but alliance-specific instead, various data processing steps must be taken place before feeding the data into our winner prediction model and alliance optimization algorithm.

\subsubsection{Distribute Match Results by Generalization}
For each match in a given year, both the ``Red'' and ``Blue'' alliances are evaluated based on how many points they earn in the default scoring activity (Traditional Scoring Low), the higher difficulty scoring activity (Traditional Scoring High), the technical scoring activity (Technical Scoring), the autonomous scoring phase (Autonomous Scoring), the endgame goal, and the number of points lost due to fouls. The number of points earned by a team in each performance indicator is calculated based on the rules of the competition in the given year. For each team, the Defense Score is defined as a negative integer denoting the total number of points that is the opposing alliance score during a match. This attribute aims to evaluate robots on how well they ``play defense'' during a match (i.e., blocking the shots of other robots, getting in their way without causing fouls, etc.).

\subsubsection{Extract Individual Robot Statistics}
The individual scores for each competing robot are computed by collecting every match in which the robot presents and calculating the mean of their alliance's performance in every performance indicator. Since the teams in each qualifying match are randomized, the given robot is the only common variable between each qualifying match. The mean of their alliance's performance in each match that the robot is part of provides an aggregate metric of overall contribution.

Scores of every robot \emph{i} are normalized for each performance indicator as $NORMSCORE = SCORE_i / SCORE_{MAX}$, where $SCORE_{MAX}$ is the maximum score earned by a robot in that performance indicator. Fouls and Defense are the exceptions since they are both negative values. Therefore, they are normalized for each performance indicator as $NORMSCORE = 1 - (SCORE_i / SCORE_{MIN})$, where $SCORE_{MIN}$ is the minimum score earned by a robot in that performance indicator.

For any alliance, its effectiveness score for each performance indicator is determined by taking the average of every participating robot's score in that performance indicator.

\subsubsection{Missing Data Handling}
The raw data from the API sometimes contains missing data due to canceled competitions, missing match data as part of a larger competition, and more. In those cases, the pipeline tries to process available matches as part of the larger competition or skips the competition if no matches are available.

\subsection{Winner Prediction Model Production}
\label{sec:model-production}
Various machine learning-driven models are tested to predict the win/loss rates of hypothetical alliances suggested by the alliance optimization algorithm. Past match data are used to evaluate the prediction model's accuracy. A multi-layer perceptron architecture is found to be the best-performing model. 

\subsubsection{Data Pre-Processing}
To train the winner prediction model, a new dataset is generated to provide the necessary team statistics of each match as well as the winner of the match. TBA's API is used to collect the necessary matches and winner data.

Since the alliance compositions are recorded for each match, the dataset of each match with winner/loser labels is expanded with each robot's seven statistics from \ref{sec:data-manipulation}. Each team's skill scores, computed in the previous pipeline, are fetched from the BigQuery table and joined to this match results table. The resulting dataset consists of 63,945 matches and 14 attributes (seven metrics for two alliances) with a binary ``red won'' as our prediction target. Fig.~\ref{fig:match_processing_pipe} shows the workflow for the processing pipeline.

Moreover, the dataset is pre-processed to contain no missing data. Once the cleaning is complete, the dataset is split into training and test sets with an 85/15 split for the model training and testing phases, respectively. The dataset contains 48,507 matches in the training set and 8,589 matches in the testing set and does not present signs of feature imbalance, providing a 50/50 split of wins for the red team and the blue team.

\begin{figure}[htbp]
\centerline{\includegraphics[width=\linewidth]{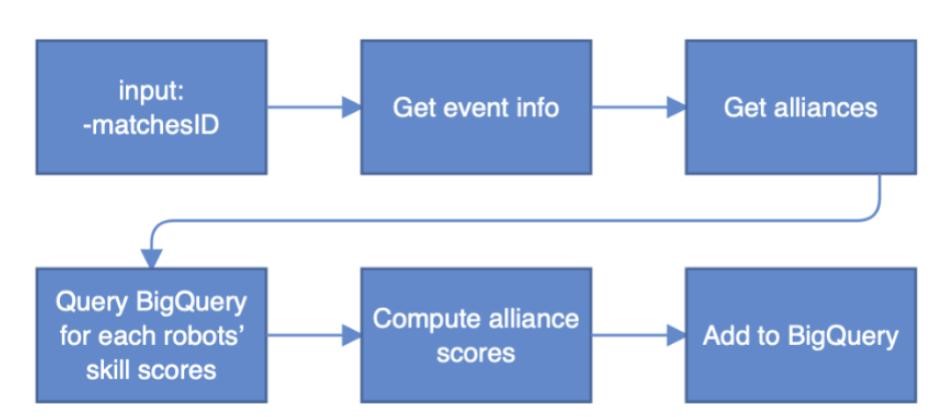}}
\caption{Match Result Processing Pipeline.}
\label{fig:match_processing_pipe}
\end{figure}

\subsubsection{Model Training}
The winner prediction model is reported following guidelines for predictive machine learning models \cite{luoGuidelinesDevelopingReporting2016}. We develop a deep neural network by using the MLPClassifier from the SKLearn library to predict the outcome of a match. The deep neural network training is optimized by using a grid search to fine-tune the hyper-parameters using the 10-fold cross-validation either over up to 500 epochs or until the convergence happens. The different combinations of hyper-parameters evaluated in the MLPClassifier are shown in Table~\ref{tab:hyperparams}.

\begin{table}[htbp]
\caption{Summary of model parameters tested to identify the best configuration for model training.}
\begin{center}
\begin{tabular}{|c|c|}
\hline
Parameter & Values Tested\\
\hline
hidden layer sizes  & (50,50,50), (50,100,50), (100,), (50, 50, 50, 50)\\
activation          & tanh, ReLU\\
solver              & SGD, Adam, L-BFGS\\
alpha               & 0.0001, 0.05\\
learning\_rate      & constant, adaptive\\
\hline
\end{tabular}
\label{tab:hyperparams}
\end{center}
\end{table}

The best combination of hyper-parameters is a (50, 50, 50, 50) hidden layer architecture, the tanh activation function, the Adam optimizer, the 0.0001 alpha value, and the constant learning rate. The winner prediction model achieves a training accuracy of 93.48\% and a testing accuracy of 84.08\% as shown in Table~\ref{tab:summary-stats}.

\begin{table}[htbp]
\caption{Summary results for the final deep neural network model. The statistics are defined as follows: Precision = True Positives / (True Positives + False Positives), Recall = True Positives / (True Positives + False Negatives), F-1 Score = 2 x [(Precision x Recall) / (Precision + Recall)], Precision = (True Positives + False Positives) / (True Positives + False Positives + True Negatives + False Negatives).}
\begin{center}
\begin{tabular}{|c|c|c|c|c|}
\hline
Outcome & Precision & Recall & F-1 Score & Accuracy\\
\hline
Blue Lost & 0.84 & 0.85 & 0.85 & \multirow{2}{*}{84.08}\\
Blue Won  & 0.84 & 0.84 & 0.84 & \\
\hline
\end{tabular}
\label{tab:summary-stats}
\end{center}
\end{table}

\subsection{Alliance Optimization}
\label{sec:alliance-optimization}
An optimization method to select alliance partners to increase overall team performance for our domain is developed. In FRC, before a competition's elimination tournament, there are a series of qualification matches to rank the teams and find the top eight robots that serve as alliance captains. These alliance captains then take turns in a snake draft to select the best teams for them to play in the elimination matches. Based on the experimental results, several options for alliance optimization are discussed. With our multi-layer perceptron model performing well in predicting which alliance may win a given match, one team-making optimization option is to use this winner prediction model to determine the average placement in the elimination bracket for an alliance upon adding a given robot to the alliance. This requires the remaining, unselected alliances to be generated procedurally, with all possible combinations of competitors being faced in the resulting playoff bracket to determine which teammate is the best to improve the team's final standing. In the real world, the team selecting their alliance partner is only given a small time to make their selection, and this solution is not computationally effective. The computational time is taken to generate all the possible playoff scenarios for each possible teammate selection and then run the winner prediction model on each scenario can be computed with the $(n-1-k)(8-k)(n-1-k)$ time complexity at a minimum, where $n$ is the number of robots in the competition and $k$ is the number of robots that are selected to be part of a team. The time complexity can be reduced to $O(n^2)$, where $n >> k$.

However, the problem can be approached more intuitively and visually based on radar plots. The suggested optimization strategy is based on the area covered in the radar plot for an alliance shown in Fig.~\ref{fig:area-maximization}. Through the development of this method, the order of variables in each compared radar plot is ensured to be consistent such that geometrically, the area calculated is comparable between alliances. This allows our tool to help generate a well-balanced team that is effective in many different fields, not just one skill area of the competition. Such a metric allows a visual and quantitative comparison of two alliances as in Fig.~\ref{fig:area-comparison}.
 
\begin{figure}[htbp]
\centerline{\includegraphics[width=0.75\linewidth]{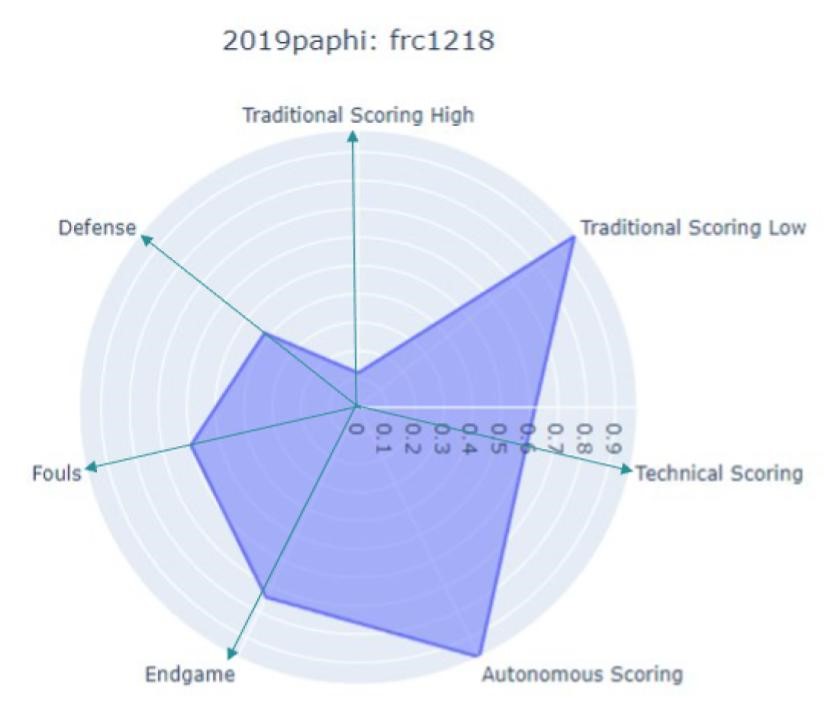}}
\caption{Maximizing Area Covered by the Radar Plot.}
\label{fig:area-maximization}
\end{figure}

\begin{figure}[htbp]
\centerline{\includegraphics[width=\linewidth]{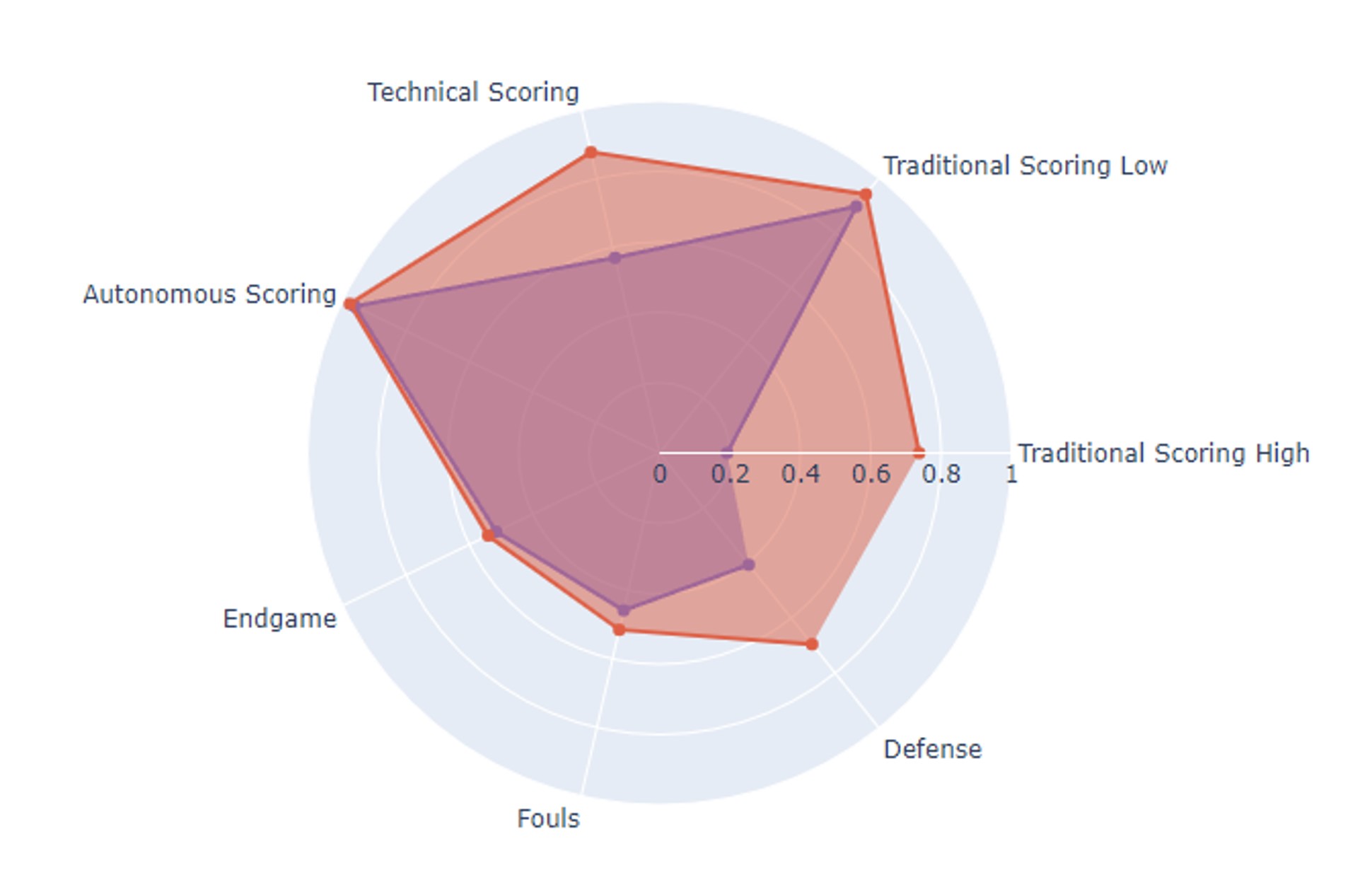}}
\caption{Two Alliances Compared Against Each Other. The Blue Alliance Is Strong at Low Scoring and Autonomous Scoring. The Red Alliance Is More Effective in All Areas of the Competition, Leading Them to the Win.}
\label{fig:area-comparison}
\end{figure}

\subsubsection{Collect Alliance Captains}
To begin this process, the top-seeded robots are acquired for the competition to be optimized. This can be done by using the rankings from TBA. The alliance captains are represented by the top eight seeded robots. These are gathered as the initial robots for each alliance to be optimized upon. This list of robot IDs is given as an argument to our optimization function.

\subsubsection{Collect Top Ranked Robots}
As a part of the alliance selection process, the top eight alliance captains can select from one another if they choose. This means that the rank 1 robot can select the rank 2 robot to become a part of their alliance. By accepting, the rank 2 robot relinquishes its position as an alliance captain, and all other robots below it in rankings move up one place accordingly (e.g., 3 moves to 2, 4 moves to 3, and so on). In this case, one of the robots outside of the initial top eight then needs to move up and become a new alliance captain. Due to this, the robots outside of the top eight are acquired from TBA's rankings. This is also given as an argument for our optimization function.

\subsubsection{Optimal Alliance Selection}
To optimize the area covered by the radar plot, a grid search is performed on the available robots to be selected, checking each robot, and calculating the area covered by their resulting alliance formed, then selecting the team that has the largest overall area. This can be calculated as an $(n-1-k)$ time complexity problem. This reduces to an $O(n)$ problem, where $n >> k$.

\subsubsection{Final Alliances}
\label{sec:final-alliances}
To create several different comparable alliance sets, we make two different versions of our optimization function.

For our ``optimize all'' function, we set up the top eight alliance captains to begin the optimization. The function then follows the snake draft, selecting the optimal alliance partner for each team consecutively until all alliances have selected their partners.

Another function allows us to act as a representative of a single alliance captain. For this function, ``optimize,'' we can enter the historical information for all the alliances that are not our representative team. When it is our representative team's turn to select in the snake draft, our optimization function kicks in and selects the best partner for the alliance. In the case where our alliance selects a team that another historical alliance selects, we, as the users, will use our best judgment to determine which teams should be selected next. With this function, we can compare how our optimized alliance compares to historical alliances from the same competition. Additionally, in an application setting, we can use this function in the alliance selection phase of a competition, and as each team is selected, we can enter it directly and have the optimized selection for our team when it is our turn to select a partner.

\section{Experimental Results and Discussions}
\label{sec:results}
The multi-layer perceptron model for the winner prediction shows an accuracy of 93.48\% when applied to the training set and 84.08\% accuracy when applied to the testing set.

Our complete methodology can accomplish alliance optimization in a competition setting in a few short steps.
\begin{enumerate}
    \item The user first gets the competition data by competition ID; e.g., \textit{2019paphi} for the 2019 Chestnut Hill Academy regional event. Data is gathered as described in Section~\ref{sec:fwk-overview} for all the robots in the competition and scores are aggregated for each robot.
    \item Next, the program automatically collects the rankings of the competition, highlighting the top eight robots. Using competition \textit{2019paphi}, the top eight robots would be, in order of rank, teams 2539, 5404, 103, 2168, 747, 3974, 1218, and 708.
    \item The users then enter their team numbers to identify themselves. For instance, if a representative from team 1218 is using our tool, they enter their team number to receive optimized alliance suggestions.
    \item Once the alliance selection begins for the competition, the user in team 1218 selects their intention to optimize only their alliance. As each alliance captain selects their partners (1218 being the 7th to choose) the user enters the chosen team numbers into the tool. In this example, team 2539 picks team 225, removing them from the selectable teams. Next, team 5404 picks team 2168. Consequently, because team 2168 is a part of the original top eight teams, the remaining teams move up, putting team 747 in the 4th place to select, then 3974, 1218, 708, and the 9th seeded team, 4342, now in the 8th position. Still being the 3rd place team, 103 selects team 747 causing another shift in the rankings, moving team 3974 to 4th place, followed by 1218, 708, 4342, and now moving team 433 into the top eight. Next up to pick is team 3974, who selects team 4342, causing yet another shift, moving 433 up to the 7th place and team 293 into the 8th. All shifts in rankings are done automatically by the tool. The user only needs to enter which teams are selected. This process is described algorithmically in Section~\ref{sec:final-alliances}.
    \item Finally, it is the user’s turn to select their first team member. With the already selected teams removed from the list, the tool goes through each of the remaining teams, as described in Section~\ref{sec:final-alliances}, calculating the total area of statistics covered for the combination of team 1218's stats and the potential alliance partners. Using this algorithm, the tool suggests the top three teams to select by potential alliance area covered. In this real-world example, team 1218 selects team 2016. However, our optimization algorithm suggests teams 5407 (to balance out team 1218’s efficient low goal scoring with some good high goal scoring), 708 (for good all-around stats), or 2016 (to double down on efficiency in the low goal).
    \item After the user selects one of the teams that is suggested (or another, sub-nominal team), the process continues, looping back in the snake draft until team 1218 is up again to choose their second alliance partner.
    \item When it is the user’s turn to select an alliance partner again, the tool goes through the same process of comparing each of the remaining teams and their resultant area until a final optimal partner is found.
\end{enumerate}

\begin{figure*}[htbp]
\centerline{\includegraphics[width=0.8\linewidth]{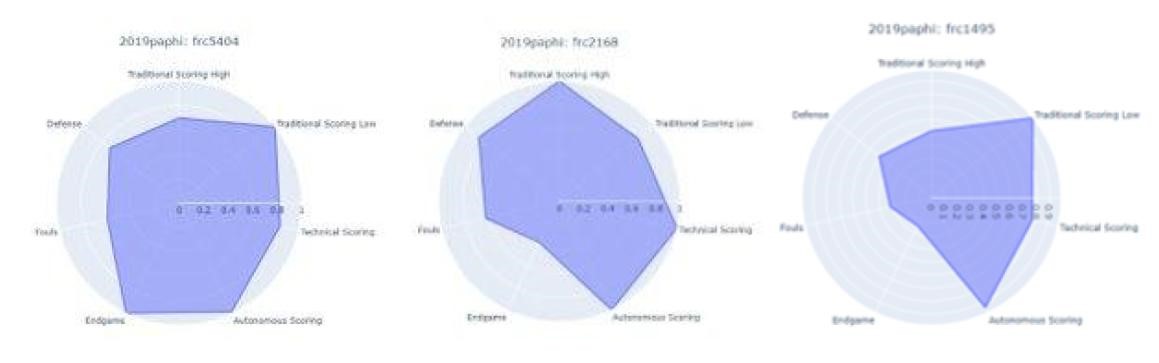}}
\caption{This Alliance Consists of teams 5404, 2168, and 1495. From Our Generalized Statistics, 5404 Is a Strong team, Particularly Good at Traditional Scoring Low, Autonomous Scoring, and End Game. Their Alliance Partner, 2168, Is Particularly Good at Traditional Scoring High and Technical Scoring but Not as Strong in the Endgame. Their Final teammate, 1495, Is Not as Strong Offensively but Could Perform Autonomously Well and Score Consistently in the Low Goal.}
\label{fig:robots-perf}
\end{figure*}

\begin{figure*}[htbp]
\centerline{\includegraphics[width=0.8\linewidth]{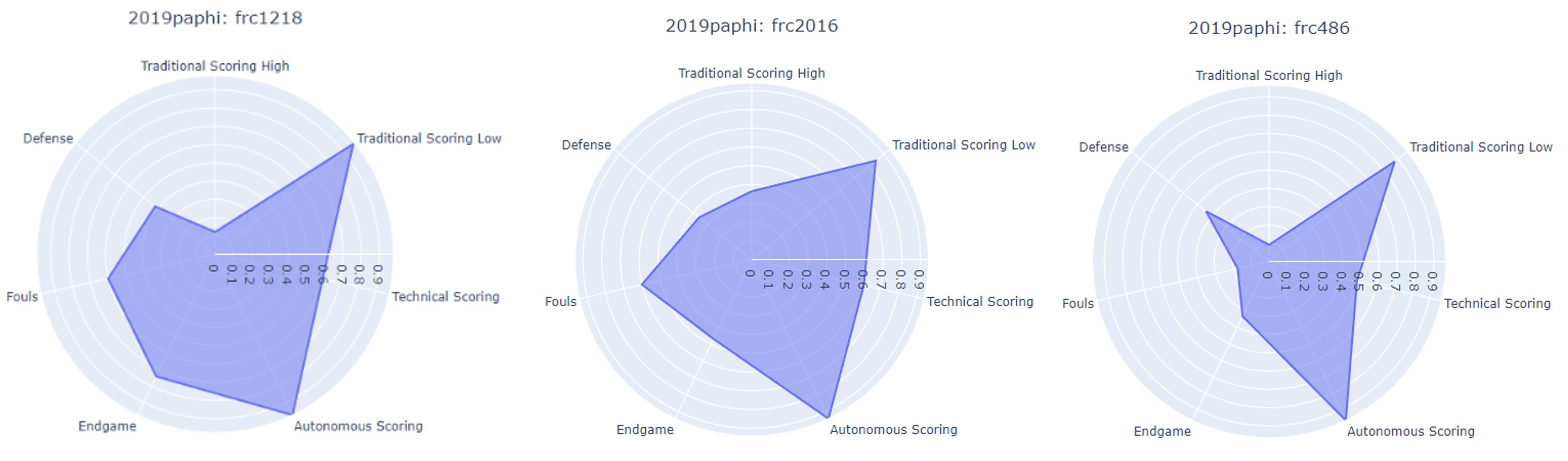}}
\caption{This Alliance Consists of teams 1218, 2016, and 486. The Alliance Captain, 1218 Is Good at Autonomous and Traditional Scoring Low With an Effective Endgame. Their teammate, 2016 has Similar Strengths but Is Not as Strong in the Endgame. Their Final Alliance Partner, 486, Is Very Specialized in Low Scoring and Autonomous, Barely Handling the Traditional Scoring High, Endgame, or Technical Scoring at All.}
\label{fig:robots-competitor-perf}
\end{figure*}

The radar plots are used to visualize robots and alliance metrics. Fig.~\ref{fig:robots-perf} is our algorithm’s evaluation of three robots competing in 2019 in the seven categories selected. For comparison, Fig.~\ref{fig:robots-competitor-perf} shows the selected set of robots that compete in the finals of a competition.
Fig.~\ref{fig:alliance-comparison} shows the calculated effectiveness of the two alliances in comparison. It can be seen that the Red team (5404, 2168, and 1495) is a stronger team overall. This translates directly to the final win for the Red Alliance taking the gold medal home.

From these preliminary visualizations, in conjunction with the implementation of our optimal alliance selection algorithm to maximize overall team stats, our framework provides a valuable strategy that teams can use to increase their chance of winning their competitions.

\begin{figure}[htbp]
\centerline{\includegraphics[width=\linewidth]{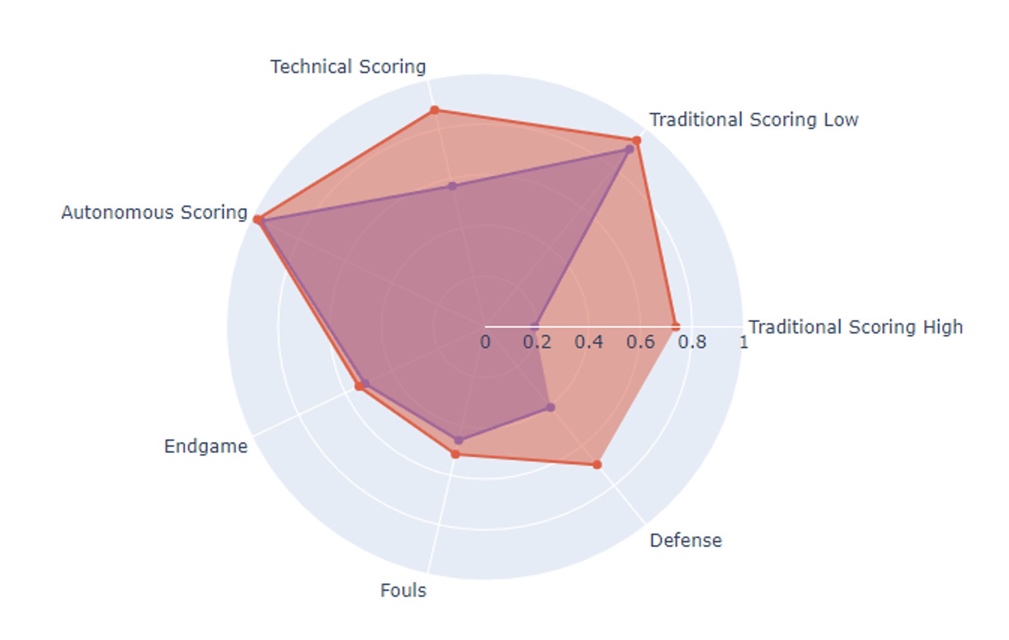}}
\caption{Radar Plot Comparing the Alliance From Fig.~\ref{fig:robots-perf} in Red, and Fig.~\ref{fig:robots-competitor-perf} in Blue.}
\label{fig:alliance-comparison}
\end{figure}

\section{Conclusions and Future Work}
\label{sec:conclusions}
The model and algorithm that we are presenting fully meet our initial goals. Our framework is able to successfully extrapolate data within our chosen domain for individual member performance from overall alliance statistics throughout the season. The alliance optimization algorithm is developed to use the extrapolated data to optimize coalitions. The deep neural network model is trained to perform winner prediction with 84.08\% accuracy. By analyzing the prediction confidence of our winner prediction model over several sets of historical and optimized alliances, it is noticed that that our team formation algorithm generates teams that are highly likely to be competitive in a challenge against other alliances.

While a popular topic of discussion amongst the FIRST Robotics community, there has yet to be any published research on the subject. Our work proposes a generalized framework that can be applied across ever-evolving seasons of data and can be adapted to different domains beyond our selected one. The importance of teammate selection is common across many different domains and our quantification methodology and selection process focus on a balanced team dynamic proved successful through our machine learning evaluation.

There is further work to be done from here. For starters, our framework does require adjustments to be made such that each category of the model fits with the rules of the most recent season. Moreover, applying computer vision to match footage can improve both alliance formation and winner prediction results. By the time of publishing, these tools have had limited development time to be usable during a competition and are an example of work still needed to improve the tool. Finally, tools to assist users in training models and running the algorithm for future years of FRC can be developed for ease of use.

\bibliographystyle{IEEEtran}
\bibliography{ref}

\end{document}